\documentclass[runningheads]{llncs}
\usepackage{amsmath,amssymb,amsfonts}
\usepackage[T1]{fontenc}
\usepackage{graphicx}
\usepackage{subfigure}
\usepackage[misc]{ifsym}

\usepackage{pifont}
\usepackage{setspace}
\usepackage{color}
\usepackage{makecell}
\usepackage{multirow}
\usepackage{booktabs}
\usepackage{url}

\usepackage[ruled, lined, linesnumbered, commentsnumbered, longend]{algorithm2e}
\SetAlCapNameFnt{\footnotesize}
\SetAlCapFnt{\footnotesize}
\usepackage{xcolor}

\SetCommentSty{mycommfont}

\begin{document}
\title{PRoA: A Probabilistic Robustness Assessment against Functional Perturbations\footnote{W. Ruan is the corresponding author.\\ W. Ruan is supported by Partnership Resource Fund (PRF) on Towards the Accountable and Explainable Learning-enabled Autonomous Robotic Systems from UK EPSRC project on Offshore Robotics for Certification of Assets (ORCA) [EP/R026173/1]. T. Zhang is supported by Exeter-CSC scholarship [202108060090].}}
\titlerunning{A Probabilistic Robustness Assessment against Functional Perturbations}
%
\author{Tianle Zhang\inst{1}\orcidID{0000-0003-4881-2406} \and
Wenjie Ruan \Letter \inst{1}\orcidID{0000-0002-8311-8738} \and
Jonathan E. Fieldsend\inst{1}\orcidID{0000-0002-0683-2583}}
\tocauthor{Tianle~Zhang (University of Exeter),
Wenjie~Ruan (University of Exeter),
Jonathan~E.~Fieldsend (University of Exeter)}
\authorrunning{T. Zhang et al.}
%

\institute{University of Exeter, Exeter EX4 4PY, UK\\
\email{\{tz294, W.Ruan, J.E.Fieldsend\}@exeter.ac.uk}}

\toctitle{PRoA: A Probabilistic Robustness Assessment against Functional Perturbations}

\maketitle              
\begin{abstract}
In safety-critical deep learning applications robustness measurement is a vital pre-deployment phase. However, existing robustness verification methods are not sufficiently practical for deploying machine learning systems in the real world. On the one hand, these methods attempt to claim that no perturbations can ``fool'' deep neural networks (DNNs), which may be too stringent in practice. On the other hand, existing works rigorously consider $L_p$ bounded additive perturbations on the pixel space, although perturbations, such as colour shifting and geometric transformations, are more practically and frequently occurring in the real world. Thus, from the practical standpoint, we present a novel and general {\it probabilistic robustness assessment method} (PRoA) based on the adaptive concentration, and it can measure the robustness of deep learning models against functional perturbations. PRoA can provide statistical guarantees on the probabilistic robustness of a model, \textit{i.e.}, the probability of failure encountered by the trained model after deployment. Our experiments demonstrate the effectiveness and flexibility of PRoA in terms of evaluating the probabilistic robustness against a broad range of functional perturbations, and PRoA can scale well to various large-scale deep neural networks compared to existing state-of-the-art baselines. For the purpose of reproducibility, we release our tool on GitHub: \url{ https://github.com/TrustAI/PRoA}.

\end{abstract}
\section{Introduction}
With the phenomenal success of Deep Neural Networks (DNNs), there is a growing and pressing need for reliable and trustworthy neural network components, particularly in safety-critical applications. Neural networks' inherent vulnerability to adversarial attacks has been receiving considerable attention from the research community~\cite{ref_Szegedy}. Numerous empirical defence approaches, including adversarial training~\cite{ref_Madry}, have been developed recently in response to diverse adversarial attacks. Such defence strategies, however, are subsequently overwhelmed by elaborate and advanced adversarial attacks~\cite{ref_Kurakin}.

Therefore, in order to construct safe and trustworthy deep learning models with a certain confidence, a challenge has emerged: \textit{how can we verify or certify our models under adversarial perturbations with guarantees}? Various earlier works have attempted to quantify the deterministic robustness of a given input $x$ concerning a specific neural network; they seek to state that no adversarial examples exist within a neighbourhood of $x$~\cite{ref_Katz_2017}. However, such safety requirements are not always satisfied and applicable in practice. For instance, as ISO/IEC Guide 51~\cite{ref_ISO} suggests, ``\textit{safety risks and dangers are unavoidable; residual risks persist even after risk reduction measures have been implemented}''. Thus, in comparison to those ensuring deterministic robustness, it is a more practical assessment of robustness to properly confine the possibility of a failure event occurring. For example, no communication networks can guarantee that no message will be lost over a wireless communication route, and messages might be lost owing to collisions or noise contamination even with proper functioning network hardware. Occasional message loss is tolerated if the occurrence chance is within an acceptable level. However, it is still unexplored for such probabilistic robustness verification.

In the meantime, the majority of existing verification methods consider a narrow threat model with additive perturbations, \textit{i.e.} adversarial examples are produced by adding slight tweaks (measured in $L_p$ distance) to every single feature of normal inputs (e.g. counterexamples are generated by adding minor changes to every single pixel in an image classification task). While the additive threat model implies that the divergence between generated adversarial instances and original instances does not surpass a modest positive constant $\epsilon$ measured by $L_p$ norm, other sorts of perturbations undetectable to humans are overlooked. For instance, cameras installed in self-driving cars may be vibrated on bumpy roads, leading to rotating or blurring photos. Resultant rotated and blurry photographs are likely to be misidentified by neural networks, even if they do not ``hoodwink'' human perception. Such risky and frequent scenarios motivate the robustness assessment against various general perturbations, e.g. geometric transformation like rotation and translation, and common corruptions.

In this paper, we propose a novel and scalable method called PRoA that can provide statistical guarantees on the probabilistic robustness of a large black-box neural network against functional threat models. Specifically, in this approach, we introduce functional perturbations, including random noise, image transformations and recolouring, which occur naturally and generally, and additive perturbation would be a specific instance in which perturbation functions add a modest adjustment to each feature of inputs. Instead of worst-case based verification, this method measures the probabilistic robustness, \textit{i.e.} accurately bounds the tolerated risk of encountering counterexamples via adaptively randomly sampling perturbations. This robustness property is more appropriate in real-world circumstances. Furthermore, the proposed method makes no assumptions about the neural network, e.g. activation functions, layers, and neurons, etc. This grants our probabilistic robustness assessment method (PRoA) the scalability to evaluate state-of-the-art and large-scale DNNs. Our main contributions are threefold as follows:
\begin{itemize}
    \item We propose a randomised algorithm-based framework for evaluating the probabilistic robustness of deep learning models using adaptive concentration inequality. This method is well-scalable and applicable to large and state-of-the-art black-box neural networks.
    
    \item The method is attack-agnostic and capable of providing a theoretical guarantee on the likelihood of encountering an adversarial example under parametric functional perturbations.
    
    \item Experimentally, we validate our certification method and demonstrate its practical applicability with different trained neural networks for various natural functional perturbations, e.g. geometric transformations, colour-shifted functions, and Gaussian blurring.
\end{itemize}

\section{Related Work}
\noindent \textbf{Reachability based approaches.} For a given input and a specified perturbation, reachability-based algorithms endeavour to determine the lower and upper bounds of the output. Thus, robustness can be evaluated by solving an output range analysis problem. Some reachability-based approaches employ layer-by-layer analysis to obtain the reachable range of outputs~\cite{ref_Li,ref_Tran,ref_Singh_Beyond,ref_Singh_An,ref_Xiang,ref_Yang,ref_Ruan}. ExactReach~\cite{ref_Xiang} estimates a DNN's reachable set as a union of polytopes by setting the outputs of each layer with Relu activation to a union of polytopes. Yang et al.~\cite{ref_Yang} present an exact reachability verification method utilising a facet-vertex incidence matrix. Additionally, another research approach is to employ global optimisation techniques to generate a reachable output interval. GeepGo~\cite{ref_Ruan} uses a global optimisation technique to find the upper and lower bounds of the outputs of Lipschitz-continuous networks. This algorithm is capable of operating on black-box DNNs. Reachability analysis can be used to address the challenge of safety verification; however, these methods often require that target networks be Lipschitz continuous over outputs, which limits their application.
\vspace{1mm}

\noindent \textbf{Constraint based approaches.} Constraint-based techniques generally transform a verification problem into a set of constraints, which can then be solved by a variety of programme solvers. In recent papers~\cite{ref_Katz_2021,ref_Amir}, Katz et al.~\cite{ref_Katz_2021} introduce an SMT-based technique called Reluplex for solving queries on DNNs with Relu activation by extending a simplex algorithm, while Amir et al.~\cite{ref_Amir}  propose another SMT-based method by splitting constraints into easier-to-solve linear constraints. For constraint-based techniques, all types of solvers can produce a deterministic answer with guarantees, \textit{i.e.}, they can either satisfy or violate robustness conditions. However, these techniques suffer from a scalability issue and need to access the internal structure and parameters of the targeted DNN (in a white-box setting).

These deterministic verification approaches might be unduly pessimistic in realistic applications since they only account for the worst scenario. In contrast, PRoA focuses on the tail probability of the average case, which is more realistic in a wide range of real-world applications, and worst-case analysis can be a special case of tail risks when we take the tail probability (0\%) of the most extreme performance into consideration.
\vspace{1mm}

\noindent \textbf{Statistical approaches.} Unlike the above deterministic verification methods, statistics-based techniques aim to quantify the likelihood of finding a counterexample. For example, random sampling has lately emerged as an effective statistical strategy for providing certified adversarial robustness, e.g. randomised smoothing~\cite{ref_Cohen,ref_Zhang_2020}, cc-cert~\cite{ref_Pautov}, and SRC~\cite{ref_Huang}, among others. Additionally, Webb et al.~\cite{ref_Webb} propose an adaptive Monte Carlo approach, \textit{i.e.} multi-level splitting, to estimate the probability of safety unsatisfiability, where failure occurs as an extremely rare occurrence in real-world circumstances. 
However, these statistics-based analyses focus on the pixel-level additive perturbations and always require assumptions upon target neural networks or distributions of input, which limits their applicability. 

In contrast, we introduce a general adversarial threat model, \textit{i.e.} functional perturbations, and PRoA aims to bound the failure chance with confidence under the functional threat model. Moreover, PRoA is able to provide rigorous robustness guarantees on black-box DNNs without any assumptions and scale to large-scale networks. 

\section{Preliminary}


\noindent \textbf{Classification program.}
Given a training set with $N$ distinct samples $S = \left\{(x_1, y_1), \dots, (x_N, y_N) \right\}$ where $x_{i} \in \mathcal{X} = \mathbb{R}^{n}$ are i.i.d. samples with dimension $n$ drawn from an unknown data distribution and $y_{i} \in \mathcal{R} = \left\{ 1, \dots, K \right\}$ are corresponding labels. We consider a deterministic neural network $f: \mathcal{X} \rightarrow [0,1]^{K}$ that maps any input to its associated output vector, and $f_k(\cdot): \mathbb{R}^{n} \rightarrow[0,1]$ is a deterministic function, representing the output confidence on label $k$. Our verification procedure solely requires blackbox assess to $f$, thus, it can obtain the corresponding output probability vector $f(x)$ for each input $x \in \mathcal{X}$.
\vspace{1mm}

\noindent \textbf{Additive Perturbation.}
Given a neural network $f$ and an input  $x \in \mathcal{X}$, an adversarial example $\widetilde{x}$ of $x$ is crafted with a slight modification to the original input such that $\underset{k \in\{1, \ldots, K\}}{\arg \max } f_{k}(\widetilde{x}) \neq \underset{k \in\{1, \ldots, K\}}{\arg \max } f_{k}(x)$; this means that the classifier assigns an incorrect label to $\widetilde{x}$ but $\widetilde{x}$ is perceptually indistinguishable from the original input $x$. Intuitively, slight perturbations $\delta \in \mathbb{R}^n$ can be added directly to $x $ to yield adversarial examples $\widetilde{x} = x + \delta$, in the meantime, a $L_p$ norm bound is normally imposed on such additive perturbations, constraining $\widetilde{x}$ to be fairly close to $x $.The relevant definition is as follows:
\begin{equation*}
    \widetilde{x} =  x + \delta \ and \ {\|\delta\|}_{p} \leq \epsilon  \ \ s.t. \ \underset{k \in\{1, \ldots, K\}}{\arg \max } f_{k}(\widetilde{x}) \neq \underset{k \in\{1, \ldots, K\}}{\arg \max } f_{k}(x) .
\end{equation*}

\noindent \textbf{Functional Perturbation.} 
Unlike the additive perturbation, a normal input $x$ is transformed using a perturbation function $\mathcal{F}: \mathcal{X} \rightarrow \mathcal{X}$ parameterised with $\theta \in \Theta$. That is to say, $x_{\mathcal{F}_\theta} =  \mathcal{F}_\theta(x)$. It is worth noting that functional perturbation allows for a substantially larger pixel-based distance, which may be imperceptible to humans as well, since the perturbed version $x_{\mathcal{F}_\theta}$ consistently preserves semantic information underlying images, such as shape, boundary, and texture. Unfortunately, such perturbations may confuse the classifier $f(\cdot)$, which is capable of outputting the proper label to an undistorted image, \textit{i.e.} $\underset{k \in\{1, \ldots, K\}}{\arg \max } f_{k}(x) \neq \underset{k \in\{1, \ldots, K\}}{\arg \max } f_{k}(x_{\mathcal{F}_\theta})$.

Prior literature on functional perturbations is surprisingly sparse. To our best knowledge, only one work involves a term \textit{functional perturbations}~\cite{ref_Laidlaw_2019}, in which a functional threat model is proposed to produce adversarial examples by employing a single function to perturb all input features simultaneously. In contrast, we introduce a flexible and generalised functional threat model by removing the constraint of global uniform changes in images. Obviously, the additive threat model is a particular case of the functional threat model, when the perturbation function $\mathcal{F}_\theta$ manipulates pixels of an image by adding slight $L_p$ bounded distortions.
\vspace{1mm}

\noindent \textbf{Verification.} 
The purpose of this paper is to verify the resilience of the classifier $f(\cdot)$ against perturbation functions $\mathcal{F}$ parameterised with $\theta \in \Theta$ while functional perturbations $\mathcal{F}_\theta$ would not change the oracle label from human perception if $\theta$ within parameter space $\Theta$, or, more precisely, to provide guarantees that the classifier $f(\cdot)$ is probabilistically robust with regard to an input $x$ when exposed to a particular functional perturbation $\mathcal{F}_ \theta$. To this end, let $k^*$ denote the ground truth class of the input sample. Assume that $\mathbb{S}_{\mathcal{F}}(x)$ is the space of all images $x_{\mathcal{F}_ \theta}$ of $x$ under perturbations induced by a perturbation function $\mathcal{F}_\theta$ and $\mathcal{P}$ is the probability measure on this space $\mathbb{S}_{\mathcal{F}}(x)$. This leads to the following robustness definitions: 
\begin{definition}[Deterministic robustness]
Let $\mathcal{F}_\theta$ be a specific perturbation function parametrized by $\theta$, and $\Theta$ denotes a parameter space of a given perturbation function. Assume that $x_{\mathcal{F}_ \theta}=\mathcal{F}_{\theta}(x)$ is the perturbed version of $x$ given $\theta \in \Theta$, and $\mathbb{S}_{\mathcal{\mathcal{F}}}(x)$ is the space of all images $x_{\mathcal{F}_ \theta}$ of $x$ under perturbation function $\mathcal{F}_\theta$. Given a K-class DNN $f$, an input $x$ and a specific perturbation function $\mathcal{F}_\theta$ with $\theta \in \Theta$, we can say that $f$ is deterministically robust w.r.t. the image $x$, \textit{i.e.} $x$ is correctly classified with probability one, if 
$$
\underset{k \in\{1, \ldots, K\}}{\arg \max } f_{k}\left(x_{\mathcal{F}}\right)=k^{*}, for \ all \  x_{\mathcal{F}} \sim \mathbb{S}_{\mathcal{\mathcal{F}}}(x).
$$
\end{definition}
\begin{definition}[Probabilistic Robustness]
Let $\mathcal{F}_\theta$ be a specific perturbation function parametrized by $\theta$, and $\Theta$ denotes a parameter space of a given perturbation function. Assume that $x_{\mathcal{F}_ \theta}=\mathcal{F}_{\theta}(x)$ is the perturbed version of $x$ given $\theta \in \Theta$, and $\mathbb{S}_{\mathcal{\mathcal{F}}}(x)$ is the space of all images $x_{\mathcal{F}_ \theta}$ of $x$ under perturbation function $\mathcal{F}_\theta$. Given a K-class DNN $f$, an input $x$, a specific perturbation function $\mathcal{F}_\theta$ with $\theta \in \Theta$, and a tolerated error rate $\tau$, the K-class DNN $f$ is said to be probabilistically robust with probability at least $1-\delta$, if 
\begin{equation}
  \mathbb{P}_{x_{\mathcal{F}} \sim \mathbb{S}_{\mathcal{F}}(x)}\left(p\left(\underset{k \in\{1, \ldots, K\}}{\arg \max } f_{k}\left(x_{\mathcal{F}}\right) \neq k^{*}\right)< \tau\right)  \geq 1-\delta. 
\end{equation}
\end{definition}
Verifying deterministic robustness has been widely studied in the context of pixel-level additive perturbations and worst-case adversarial training; however, deterministic robustness is always too stringent to hold, and deterministic robustness and probabilistic robustness are ``equivalent'' to each other when we choose $\tau =0$.

\section{Verification of Probabilistic Robustness}
We now present our proposed method, named PRoA, for verifying the probabilistic robustness of black-box classifiers against functional perturbations. A schematic overview of PRoA is illustrated in {\em Appendix \ref{app:algo}}.

\subsection{Formulating Verification Problem}
Our goal is to verify probabilistic robustness properties for a neural network classifier $f$, providing the classifier with probabilistic guarantees of its stability under functional perturbations. We formalise the robustness properties by examining substantial discrepancies of outputs w.r.t. input transformations \cite{ref_Pautov}. Next, we describe how to formalise the robustness property using both original and perturbed images.

We have a deterministic neural network $f: \mathbb{R}^{n} \rightarrow [0,1]^{K}$. Assume that a given input $x$ and its perturbed image $x_\mathcal{F}$ are assigned by $f$ with the output probability vectors ${\bf p}=f(x)$ and ${\bf p}_\mathcal{F} = f(x_{\mathcal{F}_\theta})$, respectively. Let $k^*=\arg \max \mathbf{p}$ and $\tilde{k}=\arg \max \mathbf{p}_{\mathcal{F}}$ denote the output labels assigned to original image $x$ and perturbed version $x_{\mathcal{F}_\theta}$ and $d=\frac{p_{1}-p_{2}}{2}$ be the half of the difference between two largest components of ${\bf p}$.

Then, the certain perturbations would not change the label, \textit{i.e.} $\tilde{c}=c$, if 
\begin{equation}
\left\|\mathbf{p}-\mathbf{p}_{\mathcal{F}}\right\|_{\infty}<d. 
\label{veri}
\end{equation}
where $\left\|\mathbf{p}-\mathbf{p}_{\mathcal{F}}\right\|_{\infty}=\max \left(\left|\mathbf{p}_{1}-\mathbf{p}_{\mathcal{F}_{1}}\right|, \ldots,\left|\mathbf{p}_{K}-\mathbf{p}_{\mathcal{F} K}\right|\right)$.

That means, if the maximum change caused by functional perturbations amongst all classes w.r.t. the output probability vectors, does not exceed half of the maximum difference $d$ between the two largest components of $\mathbf{p}$, the classifier will retain the category to an input $x$.  Thus, it is straightforward to provide the probabilistic guarantees that the class label assigned to an input $x$ by a classifier $f$ would not change under the transformation functions $\mathcal{F}_\theta$ by bounding the probability of the event $\left\|\mathbf{p}-\mathbf{p}_{\mathcal{F}}\right\|_{\infty}$ $<d$ occurring. 

Subsequently, we suggest applying adaptive concentration inequalities, which enable our algorithm iteratively to take more and more samples until the estimated probability of event occurrence is sufficiently accurate to be used to compute the probability satisfying Eq. \eqref{veri}. 
We establish some notation for the verification process that follows. For a random variable $Z \sim P_Z$ following any probability distribution $P_Z$, $\mu_{Z}=\mathbb{E}_{Z \sim P_{Z}}[Z]$ donates the expectation of $Z$. To fit the context of probabilistic robustness verification, we let
\begin{equation}
    Z = \bbbone \left[\left\|\mathbf{p}-\mathbf{p}_{\mathcal{F}}\right\|_{\infty}<d \right]
    \label{Z}
\end{equation} 
where $\bbbone[x]$ is an indicator function that returns 1 if $x$ is true and 0 otherwise. In this case, $\mu_Z$ represents certified stable probability of a data instance $x$ under functional transformations $\mathcal{F}_\theta$ parameterised by $\theta$, \textit{i.e.},
\begin{equation}
    \mu_Z = \mathbb{E}_{Z \sim P_{Z}}[Z] = P_{Z \sim P_{Z}}\left[Z=1 \right].
\end{equation}

\subsection{Adaptive Concentration Inequalities}
Concentration inequalities \cite{ref_Boucheron}, e.g. Chernoff inequality, Azuma's bound and Hoeffding's inequality, are fundamental statistical analytic techniques, widely applied to reliable decision-making with probabilistic guarantees. Hoeffding inequality is utilised to bound the probability of an event or the sum of bounded variables.

Let $Z$ be a random variable with distribution $P_Z$, and $Z_1, Z_2, \dots, Z_n$ are independent and identically distributed samples drawn from $P_Z$, then we can estimate $\mu_Z$, which represents the expected value of $Z$ using
\begin{equation}
    \hat{\mu}_{Z}=\frac{1}{n} \sum_{i=1}^{n} Z_{i}.
\end{equation}
Note that, regardless of the number of samples used, there must be some error $\epsilon$ between the estimated value $\hat{\mu}_{Z}$ and true expected value $\mu_{Z}$. However, we can derive high-probability bounds on this error using Hoeffding inequality \cite{ref_Hoeffding}. 
\begin{definition}[Hoeffding Inequality \cite{ref_Hoeffding}] For any $\delta>0$,
\begin{equation}
\operatorname{P}_{Z_{1}, \ldots, Z_{n} \sim P_{Z}}\left[\left|\hat{\mu}_{Z}-\mu_{Z}\right| \leq \varepsilon\right] \geq 1-\delta
\label{Hoeffding}
\end{equation}
holds for $\delta=2 e^{-2 n \varepsilon^{2}}$, equivalently, $\varepsilon=\sqrt{\frac{1}{2 n} \log \frac{2}{\delta}}$.
\end{definition}
The number of samples $n$, on the other hand, must be independent of the underlying process and determined in advance, yet in most circumstances, we generally have no idea how many samples we will need to validate the robustness specification. Consequently, we would like the number of samples used during the verification procedure to be a random variable. We decide to incorporate  {\it adaptive concentration inequality} into our algorithm, enabling our verification algorithm to take samples iteratively. Upon termination, $n$ becomes a stopping time $J$, where $J$ is a random variable, depending on the ongoing process. Then, the following adaptive Hoeffding inequality is utilised to guarantee the bound of the aforementioned probability since traditional concentration inequalities do not hold when the number of samples is stochastic.

\begin{theorem}[Adaptive Hoeffding Inequality \cite{ref_Zhao}]
Let $Z_i$ be 1/2-subgaussian random variables, and let
$\hat{\mu}_{Z}^{(n)}=\frac{1}{n} \sum_{i=1}^{n} Z_{i}$, also let $J$ be a random variable on $\mathbb{N} \cup\{\infty\}$ and let 
$\varepsilon(n)=\sqrt{\frac{a \log \left(\log _{c} n+1\right)+b}{n}}$
where $c>1, a>c/2, b>0$, and $\zeta$ is the Riemann-$\zeta$ function.
Then, we have 
\begin{equation}
  \operatorname{P}\left[J<\infty \wedge\left(\left|\hat{\mu}_{Z}^{(J)}\right| \geq \varepsilon(J)\right)\right] \leq \delta_{b} 
  \label{ori_adaptive}
\end{equation}
where $\delta_{b}=\zeta(2 a / c) e^{-2 b / c}$.
\label{adaptiveHoeffing}
\end{theorem}

\subsection{Verification Algorithm}\label{section:verification}

In this section, we will describe how to verify the probabilistic robustness of a given classifier, deriving from adaptive Hoeffding inequality. To begin, we can derive a corollary from Theorem \ref{adaptiveHoeffing}. Note that the values of $a$ and $c$ do not have a significant effect on the quality of the bound in practice \cite{ref_Zhao} and we fix $a$ and $c$ with the recommended values in \cite{ref_Zhao}, 0.6 and 1.1, respectively. 
\begin{theorem}
Given a random variable $Z$ as shown in Eq. \eqref{Z} with unknown probability distribution $P_Z$, let $\left\{Z_{i} \sim P_{\mathcal{Z}}\right\}_{i \in \mathbb{N}}$ be independent and identically distributed samples of $Z$. Let $
\hat{\mu}_{Z}^{(n)}=\frac{1}{n} \sum_{i=1}^{n} Z_{i}
$ be estimate of true value ${\mu}_{Z}$, and let stopping time $J$ be a random variable on $\mathbb{N} \cup\{\infty\}$ such that $P[J < \infty] = 1$.

Then, for a given $\delta \in \mathbb{R}_{+}$, 
\begin{equation}
\operatorname{P}\left[ \left|\hat{\mu}_{Z}^{(J)}-\mu_{Z}\right| \leq \varepsilon(\delta, J)\right] \geq 1- \delta
\label{corol}
\end{equation}
holds, where $
\varepsilon(\delta, n)=\sqrt{\frac{0.6 \cdot \log \left(\log _{1.1} n+1\right)+1.8^{-1} \cdot \log (24 / \delta)}{n}}.
$
\label{theo_adaptive}
\end{theorem}
We give a proof in {\em Appendix \ref{app:proof}}.

In the context of probabilistic robustness verification, we can certify the probabilistic robustness of a black-box neural network against functional threat models. Specifically, certified probability, $\mu_Z$, is calculated by computing the proportion of the event ($Z<d$) occurring through sampling the perturbed images surrounding an input $x$. For example, given a target neural network, we would like to verify whether there are at most $\tau$ (e.g. 1\%) adversarial examples within a specific neighbouring area around an image $x$ with greater than $1-\delta$ (e.g. 99.9\%) confidence. This means we would like to have more than 99.9\% confidence in asserting that the proportion of the adversarial examples is fewer than 1\%. 

Building upon this idea, the key of this statistical robustness verification is to prove the robustness specification of form $\mu_{Z} \geq 1-\tau$ holds. If $\mu_Z$ is quite close to $1-\tau$, then more additional samples are required to make $\epsilon$ to be small enough to ensure that $\hat{\mu}_Z$ is close to $\mu_Z$. We use a hypothesis test parameterized by a given modest probability $\tau$ of accepted violation predefined by users.  
\begin{list}{$\circ$}{}  
\item $\mathcal{H}_{0}$: The probability of robustness satisfaction $\mu \geq 1-\tau$. Thus, the classifier can be certified.
\item $\mathcal{H}_{1}$: The probability of robustness satisfaction $\mu < 1-\tau$. Thus, the classifier should not be certified.   
\end{list}
Alternatively, consider the hypothesis testing with two following conditions
\begin{equation}
\begin{aligned}
&\mathcal{H}_{0}: \hat{\mu}_{Z}+\tau-\epsilon -1 \geq 0 \\
&\mathcal{H}_{1}: \hat{\mu}_{Z}+\tau+\epsilon -1<0.
\end{aligned}   
\label{conditions}
\end{equation}
If $H_{0}$ holds, then together with $\operatorname{P}\left[ \left|\hat{\mu}_{Z}^{(J)}-\mu_{Z}\right| \leq \epsilon \right] \geq 1- \delta$, we can assert that $\mu_{Z} \geq \hat{\mu}_{Z}-\epsilon \geq 1 - \tau$ with high confidence. 
Likewise, we can conclude that 
$\mu_{Z} \leq \hat{\mu}_{Z}+\epsilon < 1 - \tau$, if $H_{1}$ holds. 

The full algorithm is summarized in Algorithm~\ref{alg:verification} in {\em Appendix \ref{app:algo}}.

\section{Experiments}
In order to evaluate the proposed method, an assessment is conducted involving various trained neural networks on public data sets CIFAR-10 and ImageNet. 

Specifically, for neural networks certified on the CIFAR-10 dataset, we have trained three neural networks based on ResNet18 architecture: a naturally trained network (plain), an adversarial trained network augmented with adversarial examples generated by $l_2$ PGD attack (AT), and a perceptual adversarial trained network (PAT) against a perceptual attack~\cite{ref_Laidlaw_2021}. In addition, four state-of-the-art neural networks, \textit{i.e.} resnetv2\_50, mobilenetv2\_100, efficientnet\_b0 and vit\_base\_patch16\_244 are introduced for ImageNet dataset; all pre-trained models are available on a PyTorch library. For our models, selected details are described in Table \ref{tab:models} in {\em Appendix \ref{app:model}}.

We provide the details about considered functional perturbations in the following subsection, and the results follow. \emph{Nota bene}, we choose $\tau=5\%$ for certifying the robustness of all models, as this is a widely accepted level in most practice. All the experiments are run on a desktop computer (i7-10700K CPU, GeForce RTX 3090 GPU). 

\subsection{Baseline setting}
To demonstrate the effectiveness and efficiency of PRoA\footnote{Our code is released via \url{https://github.com/TrustAI/PRoA}.}, it is natural to compare the estimated probability of the event, \textit{i.e.} a target model will not fail when encountering functional perturbations, obtained by PRoA with the lower limit of the corresponding confidence interval, \textit{i.e.} Agresti–Coull confidence interval (A-C CI), see {\em Appendix~\ref{app:baseline}}. 

We list the relevant existing works in Table~\ref{tab:comparison} and compare our method with these typical methods from five aspects. Specifically, DeepGo~\cite{ref_Ruan}, Reachability based~\cite{ref_Xiang}, Semantify-NN~\cite{ref_Mohapatra},
FVIM based \cite{ref_Yang} and 
CROWN \cite{ref_Zhang_2018,ref_Weng,ref_Wang} only can evaluate deterministic robustness of neural networks. Although SRC \cite{ref_Huang}, AMLS based \cite{ref_Webb}, Randomized Smoothing  \cite{ref_Cohen,ref_Zhang_2020} are able to certify probabilistic robustness, our work extensively consider models' probabilistic robustness under functional threat models.

To the best of our knowledge, there is no existing study in terms of certifying the probabilistic robustness of neural networks involving a functional threat model. Since \cite{ref_Huang} is the closest approach in spirit to our method amongst recent works, we use SRC \cite{ref_Huang} as our baseline algorithm. The proposal of SRC is to measure the probabilistic robustness of neural networks by finding the maximum perturbation radius using random sampling, and we extend it to be a baseline algorithm for computing the certified accuracy under functional perturbations.
\begin{table}[t]
\caption{Comparison with related work in different aspects.}
\centering
\scriptsize
 \setlength{\tabcolsep}{1.2mm}
 \resizebox{1\textwidth}{!}{ 
\begin{tabular}{c|c|c|c|c|c|c|c|c|c}
                         & \makecell{SRC \\ \cite{ref_Huang}}
                         & \makecell{AMLS \\ based \cite{ref_Webb}} 
                         & \makecell{Randomized \\ Smoothing  \cite{ref_Cohen,ref_Zhang_2020}} 
                         & \makecell{DeepGo \\ \cite{ref_Ruan} }
                         & \makecell{Reachability \\ based \cite{ref_Xiang}} 
                         & \makecell{Semantify-
                         \\ NN \cite{ref_Mohapatra}}
                         & \makecell{FVIM \\ based 
                         \cite{ref_Yang}}
                         & \makecell{CROWN \\ \cite{ref_Zhang_2018,ref_Weng,ref_Wang}} & PRoA \\ \hline
Deterministic Robustness & \ding{55} & \ding{55} & \ding{55} &\ding{51}  & \ding{51} & \ding{51} & \ding{51} & \ding{51} & \ding{55} \\ \hline
Probabilistic Robustness & \ding{51} & \ding{51} & \ding{51} & \ding{55} & \ding{55} & \ding{55} & \ding{55} & \ding{55} & \ding{51} \\ \hline
\makecell{Verifying Robustness on \\ Functional Perturbation}  & \ding{55} & \ding{55} & \ding{55} & \ding{55} & \ding{55} & \ding{55} & \ding{55} & \ding{55} & \ding{51}\\ \hline
Black-box Model          & \ding{51} & \ding{51} & \ding{51} & \ding{51} & \ding{55} & \ding{51} & \ding{55} & \ding{55} & \ding{51}
\end{tabular}
}
\label{tab:comparison}
\end{table}

\subsection{Considered Functional Perturbations}
PRoA is a general framework that is able to assess the robustness under any functional perturbations. In our experiments, we specifically study geometric transformation, colour-shifted function, and Gaussian blur in terms of verifying probabilistic robustness.

\vspace{1mm}

\noindent {\textbf{Gaussian Blur.}} Gaussian blurring is used to blur an image in order to reduce image noise and detail involving a Gaussian function 
\begin{equation}
G_{\theta_g}(k)=\frac{1}{\sqrt{2 \pi {\theta_g}}} \exp \left(-k^{2} /(2 {\theta_g})\right)
\end{equation}
where $\theta_g$ is the squared kernel radius. For $x \in \mathcal{X}$, we define 
\begin{equation}
\mathcal{F}_{G}{(x)} = x * G_{\theta_g}
\end{equation}
as the corresponding function parameterised by ${\theta_g}$ where $*$ denotes the convolution operator. 

\vspace{1mm}

\noindent {\textbf{Geometric Transformation.}} 
For geometric transformation, we consider three basic geometric transformations: rotation, translation and scaling. We implement the corresponding geometric functions in a unified manner using a spatial transformer block with a set of parameters of affine transformation, \textit{i.e.} $\mathcal{T}(x, \theta)$, in \cite{ref_Jaderberg}, where
\begin{equation}
\theta = \left[\begin{array}{lll}
\theta_{11} & \theta_{12} & \theta_{13} \\
\theta_{21} & \theta_{22} & \theta_{23}
\end{array}\right]
\end{equation}
is an affine matrix determined by $\theta_r$, $\theta_t$ as well as $\theta_s$.

\vspace{1mm}

\noindent {\textbf{Colour-Shifted Function.}}
Regarding colour shifting, we change the colour of images based on HSB (Hue, Saturation and Brightness) space instead of RGB space since HSB give us a more intuitive and semantic sense for understanding the perceptual effect of the colour transformation. We also consider a combination attack using brightness and contrast. 

All mathematical expressions of these functional perturbations as well as their parameter ranges are presented in {\em Appendix \ref{app:perturbation}}.
\subsection{Quantitative Results of Experiments on CIFAR-10}
To evaluate our method, we calculate the probabilistically certified accuracy of 1,000 images randomly from a test set for various functional perturbations by PRoA and SRC, in dependence on the user-defined confidence level. Furthermore, empirical robust accuracy is computed against random and grid search adversaries as well. 

To begin, we validate the effectiveness of our method over three ResNet18 models (plain, AT and PAT) trained with different training protocols against all considered functional perturbations on the CIFAR-10 dataset as mentioned previously. As a result of the experiments, we present considered perturbation functions, accompanying parameters, and quantitative results in terms of probabilistically certified accuracy (Cert. Acc), empirical accuracy (Rand.) and empirical robust accuracy (Grid) in Table \ref{tab:acc_cifar}. Clearly, the results of the proposed method align well with the validation results obtained by exhaustive search and random perturbation, and PRoA is able to achieve higher certified accuracy than SRC in almost all scenarios. Thus, the effectiveness of PRoA can be demonstrated.

\begin{figure}[!h]
\centering    
\subfigure[plain] {
\label{fig:plain}     
\includegraphics[width=0.32\columnwidth]{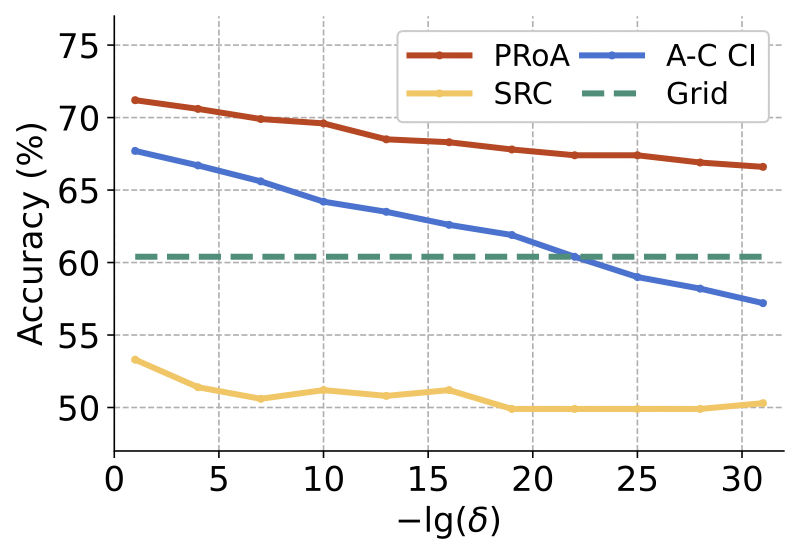}  
}   
\hspace{-6mm}
\subfigure[PAT] { 
\label{fig:pat}     
\includegraphics[width=0.32\columnwidth]{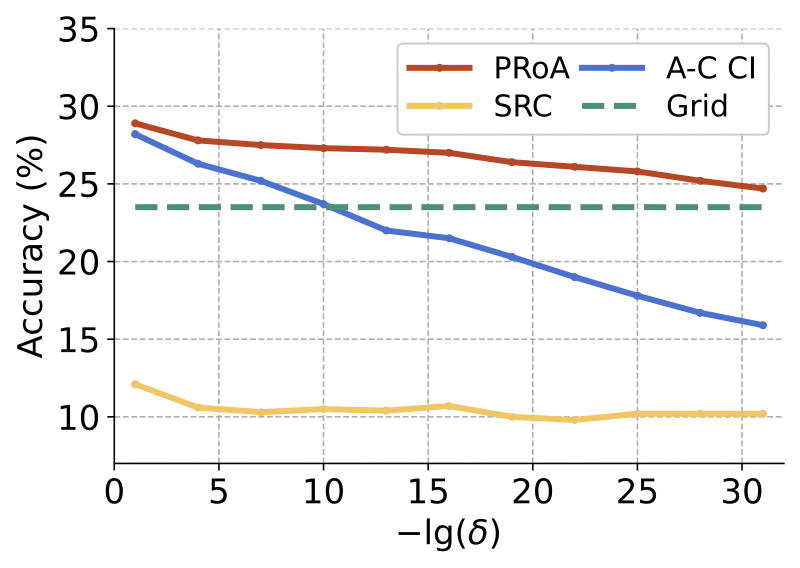}     
}   
\hspace{-6mm}
\subfigure[AT] { 
\label{fig:at}     
\includegraphics[width=0.32\columnwidth]{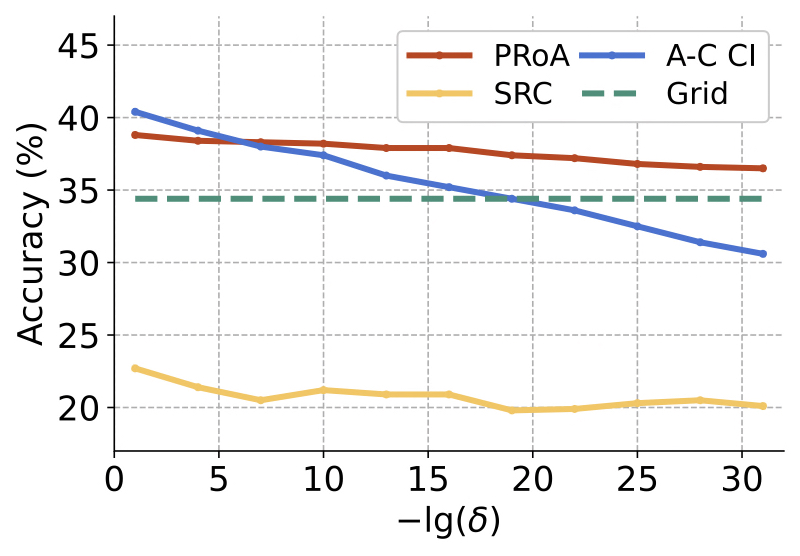}     
}   
\caption{ CIFAR-10 - An illustration of evaluating robustness of trained neural networks using PRoA, SRC and A-C CI with different confidence parameter $\delta$ against one specific perturbation, image scaling. }     
\label{fig:delta}     
\end{figure}

 \begin{table}[t]
  \centering
\scriptsize
  \caption{CIFAR-10 - Comparison of empirical robust accuracy (Grid and Rand.) and probabilistically certified accuracy (Cert. Acc) with respect to a specific model (ResNet18) with three training methods, shown in Table \ref{tab:models}. Moreover, probabilistically certified accuracy is presented with three confidence levels ($1-\delta$), low level of confidence ($\delta=10^{-4}$), middle level of confidence ($\delta=10^{-15}$) and high level of confidence ($\delta=10^{-30}$), respectively.}
  \resizebox{1\textwidth}{!}{ 
  \setstretch{1.7}
\begin{tabular}{c|c|c|c|c|ccc|ccc}
\hline
\hline
\multirow{2}{*}{Transformation}    & \multirow{2}{*}{Parameters} & \multirow{2}{*}{\makecell{Training\\ type}} & \multirow{2}{*}{Grid} & \multirow{2}{*}{Rand.} & \multicolumn{3}{c|}{\textit{SRC} Cert. Acc }  & \multicolumn{3}{c}{\textit{PRoA} Cert. Acc}                                   \\ 
\cline{6-11} 
                                  &                             &                                &                              &                                & \multicolumn{1}{c|}{$\delta=10^{-30}$} & \multicolumn{1}{c|}{$\delta=10^{-15}$} & $\delta=10^{-4}$& \multicolumn{1}{c|}{$\delta=10^{-30}$} & \multicolumn{1}{c|}{$\delta=10^{-15}$} & $\delta=10^{-4}$    \\ \midrule[0.7pt]
\multirow{3}{*}{Rotation}          & \multirow{3}{*}{$\theta_r \in [-35 ^{\circ}, 35 ^{\circ}]$}           & plain                          & 26.9\%                        &  76.8\%                                & \multicolumn{1}{c|}{24.7\% }    & \multicolumn{1}{c|}{24.8\% }   & 24.8\% & \multicolumn{1}{c|}{\textbf{30.3\%} }    & \multicolumn{1}{c|}{\textbf{31.5\%} }   & \textbf{32.0\%}     \\ \cline{3-11} 
                                  &                             & PAT                            &   16.7\%                            &            55.9\%                     & \multicolumn{1}{c|}{8.1\% }    & \multicolumn{1}{c|}{8.1\% }   & 8.1\% &   \multicolumn{1}{c|}{\textbf{10.8\%} }    & \multicolumn{1}{c|}{\textbf{12.4\%} }   &     \textbf{12.9\%}     \\ \cline{3-11} 
                                  &                        & AT                             &   16.5\%                            &           74.5\%                      & \multicolumn{1}{c|}{11.2\% }    & \multicolumn{1}{c|}{11.2\% }   & 11.2\% &     \multicolumn{1}{c|}{\textbf{14.7\%} }    & \multicolumn{1}{c|}{\textbf{15.2\%} }   &    \textbf{15.4\%}     \\ \hline
\multirow{3}{*}{Translation}       & \multirow{3}{*}{$\theta_t \in [-30 \%, 30 \%]$}           & plain                          & 62.8\%                         &          89.6\%                    & \multicolumn{1}{c|}{64.9\% }    & \multicolumn{1}{c|}{65.1\% }   & 66.6\%   & \multicolumn{1}{c|}{\textbf{77.5\%} }    & \multicolumn{1}{c|}{\textbf{78.8\%} }   & \textbf{79.4\%}     \\ \cline{3-11} 
                                  &                             & PAT                            &   50.1\%                            &      77.7\%                           & \multicolumn{1}{c|}{31.1\% }    & \multicolumn{1}{c|}{31.7\% }   & 32.4\% & \multicolumn{1}{c|}{\textbf{47.5\%} }    & \multicolumn{1}{c|}{\textbf{48.6\%} }   & \textbf{49.5\%}          \\ \cline{3-11} 
                                  &                             & AT                             &   56.5\%                          &          79.3\%                       & \multicolumn{1}{c|}{45.3\% }    & \multicolumn{1}{c|}{45.7\% }   & 46.1\% & \multicolumn{1}{c|}{\textbf{58.9\%} }    & \multicolumn{1}{c|}{\textbf{60.1\%} }   & \textbf{61.7\%}       \\ \hline 
\multirow{3}{*}{Scale}             & \multirow{3}{*}{$\theta_s \in [-70 \%, 130 \%]$}           & plain                          & 45.4 \%                         &         86.9\%   & \multicolumn{1}{c|}{48.7\% }    & \multicolumn{1}{c|}{49.0\% }   & 49.7\%                      & \multicolumn{1}{c|}{\textbf{63.3\%} }    & \multicolumn{1}{c|}{\textbf{65.2\%} }   & \textbf{67.1\%}    \\ \cline{3-11} 
                                  &                        & PAT                            &         23.5\%                      &           73.1\%   & \multicolumn{1}{c|}{8.4\% }    & \multicolumn{1}{c|}{8.7\% }   & 9.6\%                   & \multicolumn{1}{c|}{\textbf{20.9\%} }    & \multicolumn{1}{c|}{\textbf{22.8\%} }   &    \textbf{24.7\%}       \\ \cline{3-11} 
                                  &                             & AT                             &    34.4\%                           &           74.4\%  & \multicolumn{1}{c|}{19.2\% }    & \multicolumn{1}{c|}{19.4\% }   & 20.3\%                     & \multicolumn{1}{c|}{\textbf{32.4\%} }    & \multicolumn{1}{c|}{\textbf{34.5\%} }   & \textbf{35.9\%}       \\ \hline
\multirow{3}{*}{Hue}               & \multirow{3}{*}{$\theta_h \in [-\frac{\pi}{2}, \frac{\pi}{2}]$}           & plain                          & 76.9 \%                         &     89.9\%            & \multicolumn{1}{c|}{75.0\% }    & \multicolumn{1}{c|}{75.0\%}    & 75.0\%                 & \multicolumn{1}{c|}{\textbf{79.5\%} }    & \multicolumn{1}{c|}{\textbf{79.8\%} }   & \textbf{79.6\%}    \\ \cline{3-11} 
                                  &                             & PAT                            &             63.0\%                  &         77.5\%  & \multicolumn{1}{c|}{53.6\%}     & \multicolumn{1}{c|}{53.6\%}    & 53.6\%                       & \multicolumn{1}{c|}{\textbf{56.7\%} }    & \multicolumn{1}{c|}{\textbf{57.6\%} }   &\textbf{57.9\%}          \\ \cline{3-11} 
                                  &                             & AT                             &    57.6\%                           &       55.1\% & \multicolumn{1}{c|}{\textbf{54.7\%} }    & \multicolumn{1}{c|}{\textbf{54.7\%} }   & 54.7\%                         & \multicolumn{1}{c|}{54.1\% }    & \multicolumn{1}{c|}{54.5\% }   & \textbf{55.7\%}         \\ \hline
\multirow{3}{*}{Saturation}        & \multirow{3}{*}{$\theta_s \in [-30 \%, 30 \%]$}           & plain                          & 92.3 \%                         &        93.9\%         & \multicolumn{1}{c|}{95.3\% }    & \multicolumn{1}{c|}{95.3\% }   & 95.3\%                & \multicolumn{1}{c|}{\textbf{95.6\%} }    & \multicolumn{1}{c|}{\textbf{95.6\%} }   & \textbf{96.4\%}     \\ \cline{3-11} 
                                  &                             & PAT                            &             77.1\%                  &          80.8\%    & \multicolumn{1}{c|}{72.3\% }    & \multicolumn{1}{c|}{72.3\% }   & 72.4\%                     & \multicolumn{1}{c|}{\textbf{75.6\%} }    & \multicolumn{1}{c|}{\textbf{76.0\%} }   & \textbf{77.3\%}        \\ \cline{3-11} 
                                  &                             & AT                             &        74.5\%                       &             76.4\%      & \multicolumn{1}{c|}{76.8\% }    & \multicolumn{1}{c|}{77.0\% }   & 77.0\%                  & \multicolumn{1}{c|}{\textbf{79.4\%} }    & \multicolumn{1}{c|}{\textbf{79.6\%} }   & \textbf{80.3\%}     \\ \hline
\multirow{3}{*}{Brightness+Contrast}   & \multirow{3}{*}{\makecell{$\theta_b \in [-30 \%, 30 \%]$ \\
$\theta_c \in [-30 \%, 30 \%]$}  }         & plain                          & 72.6 \%                         &                92.7\%           &    \multicolumn{1}{c|}{75.5\% }    & \multicolumn{1}{c|}{75.7\% }   & 76.2\%     & \multicolumn{1}{c|}{\textbf{83.8\%} }    & \multicolumn{1}{c|}{\textbf{84.8\%} }   &   \textbf{84.1\%}      \\ \cline{3-11} 
                                  &                             & PAT                            &                   36.1\%            &         76.2\%   & \multicolumn{1}{c|}{20.5\% }    & \multicolumn{1}{c|}{20.9\% }   & 21.6\%       & \multicolumn{1}{c|}{\textbf{37.4\%} }    & \multicolumn{1}{c|}{\textbf{38.2\%} }   &  \textbf{37.0\%}                       \\ \cline{3-11} 
                                  &                             & AT                             &              31.5\%                 &      73.9\%      & \multicolumn{1}{c|}{17.8\% }    & \multicolumn{1}{c|}{17.9\% }   & 18.4\%                         & \multicolumn{1}{c|}{\textbf{34.7\%} }    & \multicolumn{1}{c|}{\textbf{38.1\%} }   & \textbf{35.7\%}     \\ \hline
\multirow{3}{*}{Gaussian Blurring} & \multirow{3}{*}{$\theta_g \in [0, 9]$}           & plain                          &        1.0\%                       &           18.1\%        & \multicolumn{1}{c|}{3.1\% }    & \multicolumn{1}{c|}{3.1\% }   & 3.3\%                & \multicolumn{1}{c|}{\textbf{3.6\%} }    & \multicolumn{1}{c|}{\textbf{3.7\%} }   &   \textbf{3.4\%}      \\ \cline{3-11} 
                                  &                             & PAT                            &   2.9\%                            &       39.7\%       & \multicolumn{1}{c|}{11.0\% }    & \multicolumn{1}{c|}{11.0\% }   & 11.0\%                       & \multicolumn{1}{c|}{\textbf{13.5\%} }    & \multicolumn{1}{c|}{\textbf{13.7\%} }   &  \textbf{12.9\%}     \\ \cline{3-11} 
                                  &                             & AT                             &                  3.7\%             &        42.9\%    & \multicolumn{1}{c|}{18.7\% }    & \multicolumn{1}{c|}{18.9\% }   & \textbf{18.9\%}                      & \multicolumn{1}{c|}{\textbf{19.2\%} }    & \multicolumn{1}{c|}{\textbf{19.3\%} }   &    18.8\%     \\ \hline \hline
\end{tabular}

  \label{tab:acc_cifar}%
 }
\end{table}%
An illustration of model verification using A-C CI, SRC and PRoA with various confidence levels ($90\%\sim1-10^{-30}$) against the picture scaling function on CIFAR-10 is depicted in Fig. \ref{fig:delta}. For instance, according to Fig. \ref{fig:plain}, we have 90\% confidence ($\delta=10^{-1}$) that this considered trained model will correctly identify roughly 71 percent of images in CIFAR-10 after a no more than 30\% image scaling with a greater than 95\% chance ($\tau=5\%$). In contrast,  we have $1-10^{-30}$ ($\delta=10^{-30}$) confidence that the proportion of images with a misclassification probability below our accepted level $5\%$ would be 67\%. Clearly, accuracy certified by PRoA reduces along with the growth of confidence, but it is not significantly changed for SRC and progressively diminishes for A-C CI. In addition, as compared with baselines, the proposed method achieves remarkable higher certified robust accuracy and a narrower gap to empirical robust accuracy along with our confidence increasing, see Fig. \ref{fig:delta}. Moreover, Grid is an approximated accuracy to the extreme case with zero tolerance ($\tau = 0$) to perturbations. However, certified accuracy with a 5\% tolerance level obtained by SRC and A-C CI always tends to be below the Grid without tolerance as the confidence level increases, which causes underestimation of the probabilistic robustness.

We apply SRC and PRoA for verifying the robustness of 500 images, which are randomly chosen from the test set on CIFAR-10. The corresponding confusion matrix is shown in Table \ref{tab:confusion}, which takes into account the cases in which the SRC outputs an ``infeasible'' status when it fails to obtain a deterministic certification result, and PRoA reaches sample limitation (set to 10,000) as a termination condition. Unsurprisingly, our method can take a certification decision in most cases when SRC returns an ``infeasible'', even though 14 images obtain a ``termination'' status due to adaptive sampling reaching sample limitation.

\vspace{2mm}

\makeatletter\def\@captype{figure}\makeatother
\hspace{-5mm}\begin{minipage}{.53\textwidth}
\centering
\includegraphics[width=1\textwidth]{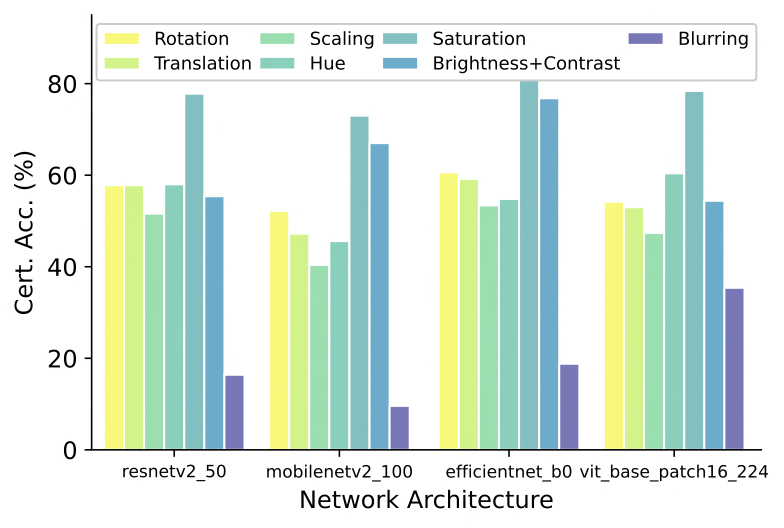}
\caption{ImageNet - Comparison of probabilistic certified accuracy with confidence level $1-10^{-4}$ , computed over the 500 randomly selected ImageNet images, amongst the models described in Table \ref{tab:models}.}
\label{fig:imagenet}
\end{minipage}
\makeatletter\def\@captype{table}\makeatother
\begin{minipage}{.45\textwidth}
  \centering
  \caption{CIFAR-10 - Confusion matrix comparing SRC and PRoA [plain model under brightness+contrast perturbation, $\delta=10^{-10}$].}
 \resizebox{1\textwidth}{!}{ 
    \begin{tabular}{ccccc}
    \hline
    \hline
          &       & \multicolumn{3}{c}{PRoA} \\
\cmidrule{3-5}          &       & Certified  & Uncertified  & Termination \\
    \midrule
    \multirow{3}[2]{*}{SRC} & Certified &   329    &  32     & 12 \\
          & Uncertified  &   36    &   9    & 0 \\
          & Infeasible &    69   &   11    & 2 \\
    \hline
    \hline
    \end{tabular}%
    }
  \label{tab:confusion}%
\end{minipage}

\subsection{Comparing probabilistic robustness across models on ImageNet}
We also use our method to analyse four large state-of-the-art neural networks against perturbation functions as mentioned earlier with 500 images randomly picked from the ImageNet test set.  Fig. \ref{fig:imagenet} demonstrates the robustness comparison of different models when subjected to diverse functional perturbations. All validation results of different models are shown as percentages in Fig. \ref{fig:imagenet}. For the `rotation' scenario, the certified accuracy of resnetv2\_50 produced by PRoA is 57.8\%, which means we have 99.99\% confidence in the claim that on average, in resnetv2\_50, 57.8\% of images will produce an adversarial example with a chance of more than 5\% in the `rotation' scenario, e.g. camera rotation. 

We also compare our algorithm to the Agresti–Coull confidence interval and SRC with a moderate confidence level, \textit{i.e.} $\delta=10^{-10}$, as shown in Table  \ref{tab:src_proa}. On the one hand, our method provides the highest certified accuracy for practically all scenarios and models; on the other hand, the average runtime of our method is comparatively longer than baselines, due to the error bounds of the estimate, which are not tight enough to make decisions and necessitate more samples. Interestingly, our algorithm takes the shortest time to certify images under a sophisticated functional perturbation, the Gaussian blurring, whereas the computation time of A-C CI and SRC increases. This is because, instead of a predetermined and decided a priori number of samples, our method terminates at any runtime $J$ depending on the ongoing process once it is capable of delivering a result, avoiding superfluous samples.

Finally, the average number of samples and the average runtime for a single image are reported in Table \ref{tab:runtime}. As one can notice, our method can be easily scaled to various SOTA network architectures, and the computation time and required samples increase reasonably with network size and complexity of perturbation function.
\vspace{-4mm}
\begin{table}[h]
  \caption{ImageNet - Comparison Agresti–Coull, SRC and PRoA [$\delta=10^{-10}$].}
\centering
 \scriptsize
 \setlength{\tabcolsep}{1.2mm}
 \resizebox{1\textwidth}{!}{ 
    \begin{tabular}{cccccccc}
    \hline
    \hline
    \multirow{2}[4]{*}{Model} & \multirow{2}[4]{*}{Perturbation} & \multicolumn{3}{c}{Certified (\%)} & \multicolumn{3}{c}{Avg. runtime (sec.)} \\
\cmidrule(l){3-5} \cmidrule(l){6-8}       &       & Agresti–Coull & SRC   & PRoA  & Agresti–Coull & SRC   & PRoA \\
    \midrule
    \multirow{7}[2]{*}{Mobilenetv2\_100} & Rotation                      &        38         &   40    &   \textbf{43}     &   \textbf{5.08}    &      5.10         &    8.35      \\
&Translation                   &        41         &   34    &    \textbf{47}    &    \textbf{5.14}    &   5.20               &    8.96      \\
&Scaling                         &       38          &   30    &    \textbf{44}    &    5.32     &   \textbf{5.07}             &     8.64   \\
&Hue                           &        40         &    \textbf{48}   &     \textbf{48}   &   5.64     &  \textbf{5.16}            &     5.19    \\
&Saturation                    &       65          &    71   &    \textbf{72}    &    5.60     &    \textbf{5.16}              &    7.26       \\
&Brightness+Contrast           &        47         &    54   &    \textbf{62}    &   5.58   &    \textbf{5.17}              &     7.03      \\
&Gaussian Blurring                      &         3        &    6   &     \textbf{8}   &   6.38    &   5.72              &     \textbf{3.89}     \\ 
    \midrule
    \multirow{7}[2]{*}{efficientnet\_b0} & Rotation                      &       46          &   47    &   \textbf{49}     &        \textbf{5.08}          &   6.25     &    6.28     \\
&Translation                   &        49         &    44   &    \textbf{57}    &         \textbf{5.14}         &    6.24    &     7.77    \\
&Scaling                         &       46           &    44   &    \textbf{51}    &        \textbf{5.32}          &    6.25    &    9.83     \\
&Hue                           &        48         &    55   &    \textbf{57}    &         \textbf{5.64}         &    6.53    &    8.69     \\
&Saturation                    &       73          &    79   &     \textbf{81}   &        \textbf{5.60}          &   6.53     &    9.83     \\
&Brightness+Contrast           &       55          &    56   &      \textbf{65}  &         \textbf{5.59}         &    6.53    &    12.37     \\
&Gaussian Blurring                      &       10          &   14    &     \textbf{17}   &        6.38          &     7.11   &     \textbf{5.61}    \\ 
    \midrule
    \multirow{7}[2]{*}{Resnetv2\_50} & Rotation                      &       51          &   46    &    \bf{54}   &       12.77           &   \textbf{9.68}     &   15.76      \\
&Translation                   &        58         &    44   &     \bf{57}   &        12.80          &    \textbf{9.56}    &    18.88    \\
&Scaling                         &       51          &   38    &    \bf{54}    &       12.80           &   \textbf{9.54}     &     17.89    \\
&Hue                           &        61         &    61   &   \bf{63}    &        13.10          &   \textbf{9.81}    &     13.04    \\
&Saturation                    &      77           &   83    & \bf{86}   &        13.15          &    \textbf{9.82}    &     16.87    \\
&Brightness+Contrast           &        39         &   32    &   \bf{40}    &       14.02           &    \textbf{9.81}     &    20.51     \\
&Gaussian Blurring                      &        15         &   14    &    \textbf{17}    &       14.16           &    10.43    &     \textbf{6.66}    \\ 
    \midrule
    \multicolumn{1}{c}{\multirow{7}[2]{*}{vit\_base\_patch16\_224}} & Rotation                      &      39           &    34   &    \textbf{41}    &       34.68           &    \textbf{33.04}    &    49.62     \\
&Translation                   &       47         &   32    &    \textbf{49}    &        34.43          &    \textbf{33.06}    &     59.18    \\
&Scaling                         &        40         &    33   &   \textbf{43}    &        34.32          &    \textbf{33.00}    &    63.21     \\
&Hue                           &        \textbf{63}         &   53    &    54    &        34.61          &    \textbf{33.31}    &    45.70     \\
&Saturation                    &        70         &    71   &    \textbf{73}    &        37.54          &    \textbf{33.31}    &     41.37    \\
&Brightness+Contrast           &       32          &    24   &    \textbf{34}    &        36.83          &    \textbf{34.19}   &    70.69     \\
&Gaussian Blurring                      &      \textbf{32}           &    28   &     30   &        35.99          &    34.88    &    \textbf{33.96}     \\ 
    \hline
    \hline
    \end{tabular}%
}
  \label{tab:src_proa}%
\end{table}

\begin{table}[!h]
\scriptsize
  \centering
  \caption{ImageNet - PRoA 
   [$\delta=10^{-10}$].}
  \resizebox{1\textwidth}{!}{ 
  \setstretch{1.15}
    \begin{tabular}{ccccc}
    \hline
    \hline
  Model & Perturbation & Avg. runtime (sec.±$std$)	& Avg. sample num.	& Certified (\%) \\
    \midrule
    \multirow{3}[2]{*}{resnetv2\_50} & Rotation  &  15.76±17.74 & 5820
  & 54 \\
        &  Brightness+Contrast &  20.51±33.05& 7930
  & 40 \\
        & Blurring  & 6.66±9.64  & 2420  & 17 \\
    \midrule
    \multirow{3}[2]{*}{mobilenetv2\_100} &  Rotation  & 8.35±13.40  & 7860
  &  43\\
        &  Brightness+Contrast & 7.03±8.23  & 6650  & 62 \\
        & Blurring  &  3.88±10.55 & 3180  & 6 \\
    \midrule
    \multirow{3}[2]{*}{efficientnet\_b0} & Rotation  &  6.28±7.28 & 4970  & 49 \\
        &  Brightness+Contrast &  12.37±13.81 & 9370  & 65 \\
        & Blurring  &  5.61±2.50 & 3790  & 17 \\
    \midrule
    \multirow{3}[2]{*}{vit\_base\_patch16\_224} & Rotation  & 49.62±80.11  &  7260
 & 41 \\
        &  Brightness+Contrast & 70.69±106.84  & 9950
  & 34 \\
        & Blurring  &  35.96±69.26 & 5020
  &  30\\
    \hline
    \hline
    \end{tabular}%
    }
  \label{tab:runtime}%
\end{table}%

\section{Conclusion}

This paper aims to certify the probabilistic robustness of a target neural network to a functional threat model with an adaptive process inspired by the Adaptive Concentration Inequalities. With PRoA, we can certify that a trained neural network is robust if the estimated probability of the failure is within a tolerance level. PRoA is dependent on the ongoing hypothesis test, avoiding a-prior sample size. The tool is scalable, efficient and generic to black-box classifiers, and it also comes with provable guarantees. 
In this paper, the hypothesis testing and adaptive sampling procedure are sequential and bring difficulty for parallelization, so one interesting future direction lies in how to further boost PRoA's efficiency, e.g., by enabling parallelization on GPUs. Another interesting future work is to bridge the gap between worst-case certification and chance-case certification. 


%
%
%
\bibliographystyle{splncs04}
%

\clearpage
\appendix

\section{Algorithm}
\label{app:algo}
\begin{figure}
    \centering
    \includegraphics[width=1\textwidth]{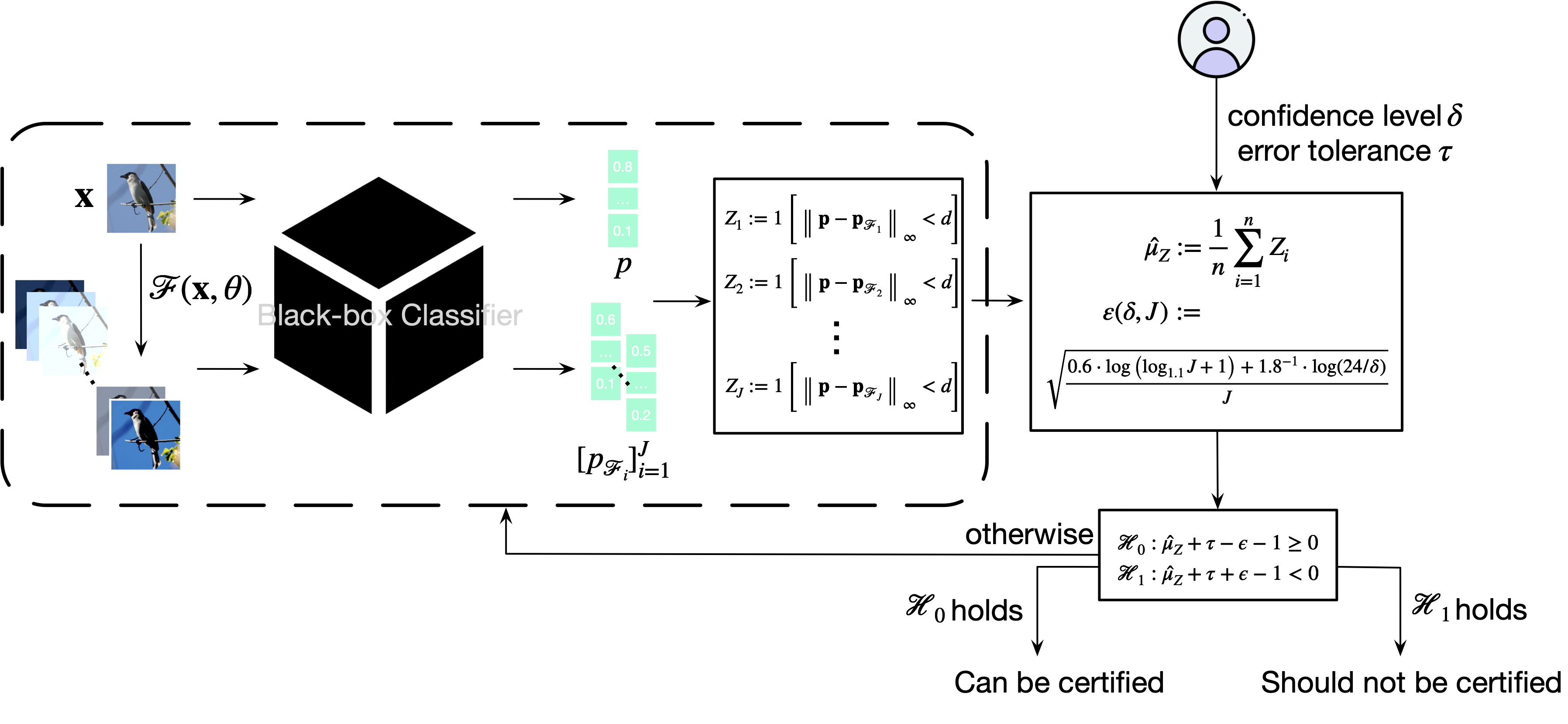}
    \caption{Illustration of Algorithm \ref{alg:verification}  verifying an single input for a black-box classifier w.r.t. a perturbation function $\mathcal{F}(x,\theta)$. }
    \label{fig:flowchart}
\end{figure}

In this section, we show the detail of the algorithm PRoA. Fig. \ref{fig:flowchart} demonstrates a schematic overview of PRoA, and the full algorithm is shown in Algorithm~\ref{alg:verification}.

\begin{algorithm}[!h]
\footnotesize
\setstretch{0.4}
\SetAlgoLined
    \caption{ Probabilistic Robustness Assessment (PRoA)}
    \label{alg:verification}
    
    \SetKwFunction{isOddNumber}{isOddNumber}
    \SetKwInOut{KwIn}{Input}
    \SetKwInOut{KwOut}{Output}

    \KwIn{ a classifier $f$, a target data point $\mathbf{x}$, functional perturbation $\mathcal{F}_\theta$, an error tolerance $\tau$, a confidence level $\delta \in \bbbr_{+}$, number of initial samples $n_0$, and maximum sample size $N_{max}$.}
    
    $\mathbf{p}=f(\mathbf{x})$ \tcp*[f]{compute output probability vector to $\mathbf{x}$}
    
    $d=\frac{\mathbf{p}[0]-\mathbf{p}[1]}{2}$ \tcp*[f]{compute the distance between top 2 classes}
    
    $hit \leftarrow y == \max{(\mathbf{p})}$
    
    \tcc{a Boolean variable indicating correctness of the classifier output for the original data $\mathbf{x}$ }
    
    set $N=n_0$
    
    \While{True}
    {
    \eIf{$J \leq N_{max}$}{
    $\mathbf{x}_{N}=$ repeat $(\mathbf{x}, N)$
    
    $\mathbf{p}_{N}=$ repeat $(\mathbf{p}, N)$
    
    $\mathbf{x}_{T}=\mathcal{F}_\theta \left(\mathbf{x}_{N}\right)$ \tcp*[f]{apply random transformation} 
    
    $\mathbf{p}_{T} \leftarrow f\left(\mathbf{x}_{T}\right)$ 
    
    \tcp{compute output probability vector of perturbed $\mathbf{x}_T$ } 
    
    $Z=\bbbone \left[\left\|\mathbf{p}-\mathbf{p}_{T}\right\|_{\infty}<d\right]$ \tcp*[f]{indicator vector of event happening }
    
    $J = J + N$
    
    $Z_J = concatenate(Z_J, Z)$ 
    
    $\hat{\mu}_Z = mean(Z_J)$
    \tcp*[f]{update estimate $\hat{\mu}_Z $ of $\mu_Z$}
    
    $\epsilon \leftarrow \varepsilon(\delta, J)$
    
    \tcc{update error tolerance of estimation w.r.t. a given confidence level $\delta$}
    
    \eIf{either condition in Eq.~\eqref{conditions} is not satisfied}
    {\KwRet{True}
    
    \tcc{termination conditions are not reached and more samples will be obtained in next iteration for more accurate estimation}
    }
    {\eIf{$H_0$ holds}{\KwOut{$Probabilistic \ robustness \ holds$}}{\KwOut{$Probabilistic \ robustness \ does \ not \ hold$}}}
    
    }{\KwOut{$Cannot \ verify \ its \ probabilistic \ robustness$}
    \tcp{termination drawing maximum samples }
    }
    }
\end{algorithm}

\section{Proof of Theorem \ref{theo_adaptive}}
\label{app:proof}

\begin{proof}
First, we give a definition about $d$-subguassian, which is a basic definition for the adaptive concentration inequalities used in \cite{ref_Zhao}.
\begin{definition}[$d$-Subguassian \cite{ref_Rivasplata}]
For any $d>0$, a random variable $Z$ is $d$-subguassian if $\mu_Z=0$ and 
$$
\mathbb{E}\left[e^{r Z}\right] \leq e^{d^{2} r^{2} / 2}
$$
holds for $\forall r \in \mathbb{R}$.
\end{definition}
Furthermore, if a distribution is bounded in a $2d$ interval, then it is a $d$-subguassian as well \cite{ref_Hoeffding}. Basically, any random variables following $d$-subguassian distribution can be scaled to be $1/2$-subguassian by $\frac{1}{2d}$. Thus, the random variable $Z-\mu_Z$ is a 1/2-subguassian distribution, implying that the probability of the bias between $\hat{\mu}_Z$ and $\mu_Z$ can be bounded by applying Theorem \ref{adaptiveHoeffing}.

Theorem \ref{theo_adaptive} follows intuitively from Theorem \ref{adaptiveHoeffing} with $a=0.6$ and $c=1.1$, because the term pertaining to the event $J<\infty$ can be omitted from Eq. \eqref{ori_adaptive} as its probability is assumed to be 1. The bound we achieved for $\mu$, Eq. \eqref{corol} is very similar to Hoeffding inequality  and Eq. \eqref{Hoeffding} can be applied to adaptively chosen stochastic stopping times, which is expected to constrain the tail probability for the threshold-crossing event, $Z<d$.
\end{proof}

\section{Experimental Details}
\label{app:experiment}
In this Appendix section, we introduce the experiment details.

\subsection{Model Details}
For a comprehensive evaluation of proposed algorithms, we adopt a set of diverse DNN models (Resnet, Mobilenet, Efficientnet and Vision Transformer). The details for model architectures are provided in Table \ref{tab:models}.
\label{app:model}
\begin{table}[h]
  \centering
  \caption{Model information.}
  \resizebox{0.8\textwidth}{!}{ 
\begin{tabular}{c|c|c|c|c}
\hline
\hline
Dataset                   & Name                    & Base Model                & Accuracy & Parameters                  \\ \midrule
\multirow{3}{*}{CIFAR-10} & plain                   & \multirow{3}{*}{Resnet18} & 99.8\%   & \multirow{3}{*}{12 Million} \\
                          & PAT                     &                           & 82.4\%   &                             \\
                          & AT                      &                           & 83.9\%   &                             \\ \hline
\multirow{4}{*}{ImageNet} & resnetv2\_50            & Resnet50                  & 80.1\%   & 26 Million                  \\
                          & mobilenetv2\_100        & Mobilenet                 & 77.3\%   & 4 Million                   \\
                          & efficientnet\_b0        & Efficientnet              & 82.3\%   & 5 Million                   \\
                          & vit\_base\_patch16\_244 & Vision Transformer        & 85.2\%   & 87 Million                  \\ \hline \hline
\end{tabular}
    }
  \label{tab:models}%
\end{table}%

\subsection{Baseline Setting }
\label{app:baseline}
Agresti–Coull confidence interval (A-C CI) is based on inverting the (large-sample) hypothesis test given in Section \ref{section:verification}. Specifically, the $100(1-\alpha)\%$ confidence interval for $p$ is
\begin{equation}
\begin{aligned}
&\text { U.L. }=\frac{\hat{p}+\frac{z_{1-\alpha / 2}^{2}}{2 n}+z_{1-\alpha / 2} \sqrt{\frac{\hat{p}(1-\hat{p})}{n}+\frac{z_{1-\alpha / 2}^{2}}{4 n^{2}}}}{1+\frac{z_{1-\alpha / 2}^{2}}{n}} \\
&\text { L.L. }=\frac{\hat{p}+\frac{z_{\alpha / 2}^{2}}{2 n}+z_{\alpha / 2} \sqrt{\frac{\hat{p}(1-\hat{p})}{n}+\frac{z_{\alpha / 2}^{2}}{4 n^{2}}}}{1+\frac{z_{\alpha / 2}^{2}}{n}} .
\end{aligned}
\end{equation}
where $z_\alpha$ is the $1-\alpha$ quantile of the standard Gaussian distribution, and $n$ is the sample size.

\subsection{Considered Functional Perturbations}
\label{app:perturbation}
Here, we detail all perturbation functions studied in this work and provide their corresponding parameters within a continuous range, while all functions are divided into geometric transformation, colour-shifted function as well as Gaussian blur function.

\noindent{\textbf{Gaussian Blur.}}
Gaussian blur is used to blur an image in order to reduce image noise and detail involving a Gaussian function 
\begin{equation}
G_{\theta_g}(k)=\frac{1}{\sqrt{2 \pi {\theta_g}}} \exp \left(-k^{2} /(2 {\theta_g})\right)
\end{equation}
where $\theta_g$ is the squared kernel radius. For $x \in \mathcal{X}$, we define 
\begin{equation}
\mathcal{F}_{G}{(x)} = x * G_{\theta_g}
\end{equation}
as the corresponding function parameterised by ${\theta_g}$ where $*$ denotes the convolution operator. The blur factor $\theta_g$ is constrained in (0, 9) in all our experiments.

\noindent{\textbf{Geometric Transformation.}}
We consider three typical geometric transformations: rotation, translation and scaling. We implement the corresponding geometric functions using spatial transformer networks with a set of parameters of affine transformation in a unified manner.
\begin{itemize}
    \item [$\bullet$] {\it Rotation}  \ \ Rotating image around the centre in an angle can cause misclassification by model, and the function of rotation is parameterised by the rotate angle $\theta_r \in \mathbb{R}$. In this case, we define $\theta_r \in [-35^{\circ}, 35^{\circ}]$ for CIFAR-10 dataset while $\theta_r \in [-30^{\circ}, 30^{\circ}]$ for ImageNet dataset.
    
    \item [$\bullet$] {\it Translation}  \ \ An image is shifted in coordinate in both vertical and horizontal directions, and the associated function is parameterised by a 2-D vector $\theta_t \in \mathbb{R}^2$. In our experiments, we specify $\theta_t \in [-30 \% \times w, 30 \% \times w]$, where $w$ is the width (equal to height) of the picture, \textit{i.e.} translation does not exceed 30\% of the width and height of images.

    \item [$\bullet$] {\it Scaling}  \ \ Resizes images, its corresponding function is controlled by a scale rate $\theta_s \in \mathbb{R}$. We set the scale rate $\theta_s$ within the interval [0.7, 1.3] for ImagNet and CIFAR-10, which means we modify image size by no more than 30\% for images from ImageNet and CIFAR-10.
\end{itemize}
For the case of geometric transformations, we implement the associated functions, $\mathcal{F}_R(x, \theta_r)$, $\mathcal{F}_T(x, \theta_t)$ and $\mathcal{F}_S(x, \theta_s)$ in a unified manner using a spatial transformer block, $\mathcal{T}(x, \theta)$, in \cite{ref_Jaderberg}, where
\begin{equation}
\theta = \left[\begin{array}{lll}
\theta_{11} & \theta_{12} & \theta_{13} \\
\theta_{21} & \theta_{22} & \theta_{23}
\end{array}\right]
\end{equation}
is an affine matrix determined by $\theta_r$, $\theta_t$ as well as $\theta_s$.

\noindent{\textbf{Colour-Shifted Function.}}
We change the colour of images based on HSB (Hue, Saturation and Brightness) space instead of RGB space, since HSB give us a more intuitive and semantic sense for understanding the perceptual effect of the colour transformation. We also consider a combination attack using brightness and contrast. Thus, we can define the following functions:
\begin{itemize}
    \item [$\bullet$] {\it Hue}  \ \ Hue refers to every variety of colours of the visual spectrum, \textit{i.e.} red, yellow, green and blue, etc., the scale of which is always represented as a colour wheel between $-\pi$ and $\pi$. In our experiments, we use $\theta_h \in [-\frac{\pi}{3}, \frac{\pi}{3}]$ bounding the parameter of hue function,
    \begin{equation}
    \mathcal{F}_H(x^h, \theta_h) = (x^h + \theta_h) \bmod (2 \pi) 
    \end{equation}
    
    \item [$\bullet$] {\it Saturation}  \ \  Colour saturation determines the intensity of colour in an image, that is, If the saturation value $x^s$ increases, the colour becomes more pure; if $x^s$ decreases, the colour appears to more gray. In our experiments, we certify the whole range of saturation factor, \textit{i.e.} $\theta_s \in [-0.5, 0.5]$ and the saturation function is
    \begin{equation}
    \mathcal{F}_S(x^s, \theta_s) =\min(\max(0, (1+\theta^s) \cdot x^s), 1)
    \end{equation}

    \item [$\bullet$] {\it Brightness and Contrast}  \ \ We first perturb the brightness of an image by adding a constant value $\theta_b$ to each pixel, and then change the image contrast by an pixel-wise multiplication with a positive contrast parameter $\theta_c$. Thus, we can define the brightness and contrast function $\mathcal{F}_{BC}: \mathcal{X} \times \mathbb{R}^{2} \rightarrow \mathcal{X}$ as  
    \begin{equation}
    \mathcal{F}_{BC}{(x, \theta)} = min(max((1+\theta_c)\cdot x + \theta_b, 0), 255), \ \ \ \ \theta=(\theta_b, \theta_c)^T
    \end{equation}
    In our experiments, we use $\theta_b \in (-30\%, 30\%)$ and $\theta_c \in (-30\%, 30\%)$, \textit{i.e.} we only alter brightness and contrast of images by less than 30\%.
\end{itemize}

\end{document}